\documentclass[letterpaper]{article} 
\usepackage{aaai2026}  
\usepackage{times}  
\usepackage{helvet}  
\usepackage{courier}  
\usepackage[hyphens]{url}  
\usepackage{graphicx} 
\usepackage{natbib}  
\usepackage{caption} 
\usepackage{algorithm}
\usepackage{algorithmic}
\usepackage{amsmath}
\usepackage{amssymb}
\usepackage{bm} 
\usepackage{pifont}
\usepackage{booktabs}
\usepackage{adjustbox}
\usepackage{multirow}
\usepackage{xcolor}
\usepackage[table]{xcolor}
\definecolor{Gray}{RGB}{240,240,240}

\usepackage{newfloat}
\usepackage{listings}
\DeclareCaptionStyle{ruled}{labelfont=normalfont,labelsep=colon,strut=off} 
\floatstyle{ruled}
\newfloat{listing}{tb}{lst}{}
\floatname{listing}{Listing}
\title{Explore How to Inject Beneficial Noise in MLLMs}
\author{
    Ruishu Zhu\textsuperscript{\rm 1,2},
    Sida Huang\textsuperscript{\rm 1,2},
    Ziheng Jiao\textsuperscript{\rm 3},
    Hongyuan Zhang\textsuperscript{\rm 2, 4}\thanks{Corresponding author.}
}
\affiliations{
    \textsuperscript{\rm 1}School of Artificial Intelligence, OPtics and ElectroNics (iOPEN), Northwestern Polytechnical University \\
    \textsuperscript{\rm 2}Institute of Artificial Intelligence (TeleAI), China Telecom\\
    \textsuperscript{\rm 3}HuaWei Technologies Co., Ltd.\\
    \textsuperscript{\rm 4}The University of Hong Kong\\
    \{zhuruishu0848, sidahuang2001\}@gmail.com, jzh9830@163.com, hyzhang98@gmail.com
}

\usepackage{bibentry}

\begin{document}

\maketitle

\begin{abstract}
 Multimodal Large Language Models (MLLMs) have played an increasingly important role in multimodal intelligence. However, the existing fine-tuning methods often ignore cross-modal heterogeneity, limiting their full potential. In this work, we propose a novel fine-tuning strategy by injecting beneficial random noise, which outperforms previous methods and even surpasses full fine-tuning, with minimal additional parameters. The proposed Multimodal Noise Generator (MuNG) enables efficient modality fine-tuning by injecting customized noise into the frozen MLLMs. Specifically, we reformulate the reasoning process of MLLMs from a variational inference perspective, upon which we design a multimodal noise generator that dynamically analyzes cross-modal relationships in image-text pairs to generate task-adaptive beneficial noise. Injecting this type of noise into the MLLMs effectively suppresses irrelevant semantic components, leading to significantly improved cross-modal representation alignment and enhanced performance on downstream tasks. Experiments on two mainstream MLLMs, QwenVL and LLaVA, demonstrate that our method surpasses full-parameter fine-tuning and other existing fine-tuning approaches, while requiring adjustments to only about $1\sim2\%$ additional parameters. The relevant code is uploaded in the supplementary. 
\end{abstract}

\begin{links}
    \link{Code}{https://github.com/zhuruishu0848/MuNG}
\end{links}


\section{Introduction}\label{sec:intro}
In recent years, Large Language Models (LLMs)~\cite{touvron2023llama,achiam2023gpt,qwen2.5,deepseekai2025deepseekr1incentivizingreasoningcapability} have demonstrated impressive capabilities, successfully addressing various complex tasks. Building upon this strength, cutting-edge works such as LLaVA~\cite{liu2023visual}, Qwen2.5-VL~\cite{Qwen2.5-VL}, and InternVL~\cite{chen2024internvl} have begun exploring the synergy between vision and language modalities. Integrating vision and language allows models to transcend textual limits, gain visual understanding, and produce fluent visual descriptions. This cross-modal fusion is driving a paradigm shift in large models from single-modal understanding to multi-modal interaction, while injecting new developmental momentum into multiple AI frontiers.

However, current mainstream models still exhibit notable limitations in tasks involving spatial relationship understanding~\cite{rahmanzadehgervi2024vision,hudson2019gqa}, hallucination suppression~\cite{tong2024eyes,li2023evaluating}, and over-reliance on textual information~\cite{rahmanzadehgervi2024vision}. To address these issues, full-parameter fine-tuning (FT) is commonly employed, which involves updating all model parameters. However, for large-scale models, this approach entails substantial computational overhead and may lead to overfitting, especially when fine-tuning data is limited, thereby compromising the model-acquired general and cross-modal knowledge.


To improve efficiency, researchers initially attempted to adopt Parameter-Efficient Fine-Tuning (PEFT)~\cite{houlsby2019parameter} techniques from LLMs, updating or adding minimal parameters to the LLM Decoder. For example, LoRA~\cite{hu2022lora} achieves substantial parameter compression through low-rank matrix injection, while Adapter employs a bottleneck structure to update only a small number of parameters. In addition, Visual Prompt Tuning (VPT)~\cite{jia2022visual} approaches the problem from the visual modality perspective by introducing a small set of learnable prompts in the input space after visual embedding layers to adapt to downstream tasks. However, these fine-tuning methods remain fundamentally rooted in single-modal optimization paradigms, neglecting the need for vision-language co-optimization. Consequently, models find it difficult to efficiently adapt to the distribution shifts and alignment requirements of downstream task data. \textbf{Thus, there is an urgent need to establish fine-tuning methodologies tailored to multi-modal characteristics.}

In this work, we adopt a distinct approach. Instead of fine-tuning the LLM Decoder or unimodal Encoders, we modify the input content fed into the LLM Decoder. Motivated by the theoretical insights of positive-incentive noise~\cite{Li2022PositiveIncentiveN, VPN, PiNI, PiNDA, PiNGDA, MIN}, we present a novel, lightweight, and effective fine-tuning approach aimed at improving the multimodal understanding capabilities of MLLMs.

Our method introduces only a small set of learnable parameters during alignment to model noise distributions, while freezing the entire pretrained Encoder and Decoder backbones during training. The noise injection process enables the data to learn more generalized representations. During inference, these additional parameters are solely used to generate beneficial noise, which is added to the LLM Decoder’s input.

\begin{figure*}[t]
  \centering
  {\includegraphics[width=1\textwidth]{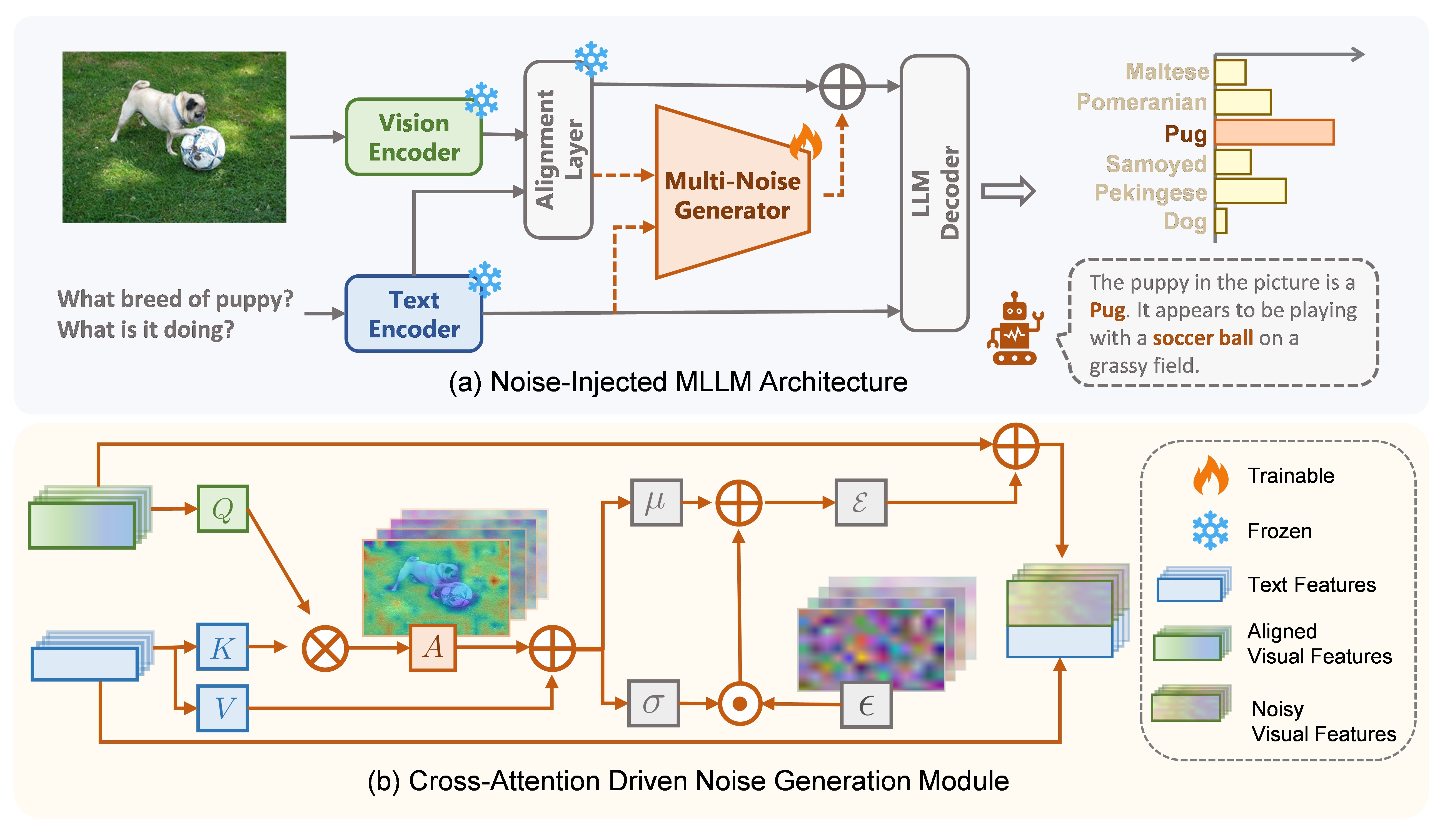}}
  \caption{Pipeline of MuNG. \textbf{(a) The overall framework of noise injection in MLLMs.} The proposed MuNG is inserted between the feature alignment layer and the LLM decoder, injecting task-adaptive beneficial noise into the visual representations.
\textbf{(b) The architecture of the multimodal noise generator based on cross-attention.} A random signal $\epsilon$ is sampled from a standard normal distribution and combined with the mean and variance obtained via cross-attention to generate the final noise.
In the figure, $\odot$ denotes the Hadamard product, $\oplus$ denotes matrix or vector addition, and $\otimes$ denotes matrix multiplication.}
  \label{fig:Pipeline}
\end{figure*}

Our contributions can be summarized as follows:

\begin{enumerate}
    \item We propose \textbf{Mul}timodal \textbf{N}oise \textbf{G}enerator (MuNG), the \textbf{first method} to leverage multimodal information for \textbf{generating beneficial noise to fine-tune MLLMs}. By reformulating the inference process of MLLMs from a variational inference perspective.
    
    \item MuNG injects beneficial noise into model, \textbf{effectively suppressing irrelevant semantics} and enhancing coherent cross-modal representations. Our method introduces \textbf{only around $1\%$ additional parameters}, enabling a highly parameter-efficient fine-tuning strategy.
    
    \item Experiments demonstrate that our approach outperforms previous parameter-efficient methods and \textbf{even matches or exceeds full fine-tuning performance} across several benchmarks. Experiments validate our method’s efficiency and generalizability.

\end{enumerate}

\section{Related Work}
\label{gen_inst}

\subsection{Multimodal Large Language Models}
In recent years, Large Language Models (LLMs) have demonstrated impressive capabilities in the text modality, successfully addressing various complex tasks. Building upon this strength, cutting-edge works such as LLaVA~\cite{liu2023visual}, Qwen-VL~\cite{bai2023qwen}, GPT-4o~\cite{hurst2024gpt}, and InternVL~\cite{chen2024internvl} have begun exploring the synergy between vision and language modalities—as the two most critical modalities in the real world, their interconnection not only enables language models to break through textual limitations and acquire visual perception abilities but also facilitates the generation of more vivid and fluent linguistic descriptions of visual scenes. This cross-modal fusion is driving a paradigm shift in large models from single-modal understanding to multi-modal interaction, while injecting new developmental momentum into multiple AI frontiers, including embodied intelligence. Meanwhile, text-to-image generation models have shown remarkable progress in visual synthesis from textual inputs~\cite{LDM, adv-CPG, AVEdit, NFIG}.

\subsection{PEFT Methods for Large Models }
To reduce the computational overhead and memory footprint of fine-tuning large-scale models, various Parameter-Efficient Fine-Tuning (PEFT) methods have been proposed. Among them, Adapter~\cite{houlsby2019parameter} , BitFit ~\cite{zaken2021bitfit}, and LoRA ~\cite{hu2022lora}have demonstrated strong empirical performance across a wide range of downstream tasks. Adapter-based methods insert small trainable modules into model layers while keeping most parameters frozen. BitFit further simplifies adaptation by tuning only bias terms, yet still achieves competitive results, particularly in text classification. LoRA introduces low-rank matrices into attention layers, significantly reducing trainable parameters without sacrificing performance. These methods reflect a growing trend: achieving high effectiveness with minimal parameter updates—ideal for resource-constrained or multi-task settings.
In multimodal settings, some fusion modules—such as the Multimodal Bottleneck Transformer (MBT)~\cite{nagrani2021attention}, which inserts fusion bottleneck tokens to enable cross-modal interaction—can also act as efficient tuning strategies by enhancing modality integration.

\subsection{Difference with MLLM Adversarial Attacks }
Injecting adversarial noise into models is a common white-box attack technique. Some studies have extended classical adversarial attack methods from image classification tasks—such as FGSM ~\cite{goodfellow2014explaining} and PGD ~\cite{madry2017towards}—to Multimodal Large Language Models ~\cite{zhao2023evaluating, carlini2023aligned, qi2024visual, wang2024white}. These methods generate adversarial noise using gradient information and introduce imperceptible perturbations to the input data, thereby inducing MLLMs to produce incorrect or even unsafe outputs.In contrast, the noise introduced by our method belongs to the category of positive incentive noise. Rather than attacking the model, the goal is to simplify the task itself and guide the model toward generating more accurate outputs. This approach emphasizes enhancing model performance rather than launching untargeted attacks.

\section{Method}
\label{headings}


\subsection{Definition of $\pi$-noise in Multimodal Task}
Current mainstream vision-language large models typically consist of modality encoders, a multimodal fusion layer, and a LLM decoder. First, the visual and linguistic inputs, are independently processed by their respective encoders to obtain visual features and language features. These features are then integrated through a multimodal fusion layer to enable cross-modal interaction, resulting in the visual features $X_{V}$ and language features $X_{L}$. And fed them into the LLM Decoder to produce the final linguistic output $A$. According to the theory established by ~\cite{Li2022PositiveIncentiveN}, the complexity of a VQA task can be defined as
\begin{equation}
\begin{aligned}
  H\left( \mathcal{T} \right) 
  &= \int_{\mathcal{X}_V} \int_{\mathcal{X}_L} p(X_V, X_L) \\
     &\left( \int_{\mathcal{A}} -p(A | X_V, X_L) \log p(A | X_V, X_L)  dA \right) dX_L dX_V \\
  &= \mathbb{E}_{p\left( A, X_{V}, X_{L} \right)} \left[ -\log p\left( A | X_{V}, X_{L} \right) \right],\\
\end{aligned}
\end{equation}
where $A$ denotes the expected linguistic output, $X_{V}$ and $X_{L}$ represent the input visual and language features, and $\mathcal{D}_{X_{V}}, \mathcal{D}_{X_{L}}$ are the data distributions over $X_{V}$ and $X_{L}$ respectively.

If a noise 
$I\left( \mathcal{T}, \mathcal{E} \right) = H\left( \mathcal{T} \right) - H\left( \mathcal{T} \mid \mathcal{E} \right) > 0 \Leftrightarrow H\left( \mathcal{T} \right) > H\left( \mathcal{T} \mid \mathcal{E} \right)$
then injecting this noise into MLLM can reduce task uncertainty and, in turn, simplify the task complexity. 

Since $H(\mathcal{T})$ remains constant for a given MLLM, maximizing $I(\mathcal{T}, \mathcal{E})$ is equivalent to minimizing the conditional entropy $H(\mathcal{T} \mid \mathcal{E})$, which can be formulated as
\begin{equation}
\begin{aligned}
	H\left( \mathcal{T} | \mathcal{E} \right) = \mathbb{E}_{p\left( A, X_{V}, X_{L}, \mathcal{E} \right)}
	\left[ -\log p\left( A | X_{V}, X_{L}, \mathcal{E} \right) \right].
\end{aligned}
\end{equation}

\subsection{Variational Approximation}
Due to the intractability of directly computing $p\left( A | X_{V}, X_{L}, \varepsilon \right)$, we adopt variational inference techniques. Leveraging the non-negativity of the KL divergence,
\begin{equation}
	\begin{aligned}
		KL(p||q) \ge 0 \Leftrightarrow \mathbb{E}_{p(x)}[\log p(x)] \ge \mathbb{E}_{p(x)}[\log q(x)],
	\end{aligned}
\end{equation}
we can approximate $p\left( A \mid X_{V}, X_{L}, \mathcal{E} \right)$ and derive a variational upper bound,
\begin{equation}
	\begin{aligned}
		L = \mathbb{E}_{p\left( A, X_{V}, X_{L}, \mathcal{E} \right)} \left[ -\log q\left( A | X_{V}, X_{L}, \mathcal{E} \right) \right] \ge H\left( \mathcal{T} | \mathcal{E} \right).
	\end{aligned}
\end{equation}
However, minimizing this variational upper bound is still challenging. Therefore, we employ Monte Carlo sampling from the data distributions $\mathcal{D}_{X_{V}}$ and $\mathcal{D}_{X_{L}}$ to obtain image-question-answer triplets $(X_{V_i}, X_{L_i}, A_i)$. By approximating the expectation using sampled triplets, we derive the following loss function:
\begin{equation}
	\begin{aligned}
		L \approx \frac{1}{n} \sum_{i=1}^{n} \mathbb{E}_{p\left( \mathcal{E} | A_i, X_{V_i}, X_{L_i} \right)} \left[ -\log q\left( A_i | X_{V_i}, X_{L_i}, \mathcal{E} \right) \right].
	\end{aligned}
\end{equation}
Assuming a Gaussian distribution, we use a learnable function to approximate the mean $\mu$ and variance $\sigma$,
\begin{equation}
	\begin{aligned}
		\mu, \sigma = f_{\theta}(A_i, X_{V_i}, X_{L_i}),
	\end{aligned}
\end{equation}
 $\theta$ denotes the learnable parameters. To ensure that gradients can be backpropagated through the sampling process, we apply the reparameterization trick by introducing an auxiliary variable $\epsilon \sim \mathcal{N}(0, 1)$, which decouples the randomness from the network’s parameters. This ensures that the randomness in $\mathcal{E}$ is compatible with gradient-based optimization, while $\mu$ and $\sigma$ remain differentiable,
\begin{equation}
	\begin{aligned}
		\mathcal{E} = G_{\theta}(\epsilon, A_i, X_{V_i}, X_{L_i}) = \sigma \cdot \epsilon + \mu.
	\end{aligned}
     \label{eq:Reparameterization}
\end{equation}
The variational upper bound can thus be approximated as
\begin{equation}
	\begin{aligned}
		L \approx & \frac{1}{n} \sum_{i=1}^{n} \mathbb{E}_{p\left( \mathcal{E} | A_i, X_{V_i}, X_{L_i} \right)} \\
        &\big[ -\log q\left( A_i | X_{V_i}, X_{L_i}, G_{\theta}(\epsilon, A_i, X_{V_i}, X_{L_i}) \right) \big].
	\end{aligned}
\end{equation}
For each triplet $(A_i, X_{V_i}, X_{L_i})$, we draw $m$ samples of $\epsilon$ to estimate the final training loss,
\begin{equation}
	\begin{aligned}
		L & \approx \frac{1}{n \cdot m} \sum_{i=1}^{n} \sum_{j=1}^{m} \\
        &\big[ -\log q\left( A_i \mid X_{V_i}, X_{L_i}, G_{\theta}(\epsilon_{ij}, A_i, X_{V_i}, X_{L_i}) \right) \big].
	\end{aligned}
    \label{eq:loss}
\end{equation}

\subsection{Specific Implementation}
Based on the training loss function derived in Eq.~\eqref{eq:loss} and the architectural characteristics of MLLMs, our approach can be implemented in two stages. First, we generate noise $\mathcal{E} = G_{\theta}(\epsilon, A_i, X_{V_i}, X_{L_i})$ using multimodal information, including visual features, textual features, and the target output. Then, the generated noise $\mathcal{E}$ is injected into the MLLM to produce the final textual output.The specific architecture is shown in Figure~\ref{fig:Pipeline}.

\subsubsection{Multimodal Noise Generator.} Based on the formulation $\mathcal{E} = G_{\theta}(\epsilon, A_i, X_{V_i}, X_{L_i})$, we design a multimodal noise generator. During inference, we use $\mathcal{E} = G_{\theta}(\epsilon, X_{V_i}, X_{L_i})$.
Specifically, we feed the visual features $X_V$, textual features $X_L$, and target textual output $A$ into the noise generation module, which learns the distribution parameters $\mu$ and $\log(\sigma)$. Using the reparameterization trick, the noise $\mathcal{E}$ is then generated following Eq.~\eqref{eq:Reparameterization}.
During training, we concatenate the question and target answer as the model input, but compute the loss only over the answer portion, ignoring the prediction error of the question. This strategy resembles the supervised fine-tuning process of LLMs, allowing us to reuse the original LLM Decoder structure without modification. The function $f_{\theta}(A_i, X_{V_i}, X_{L_i})$ can be implemented using various n
eural architectures, such as MLPs or cross-attention modules. We provide a comprehensive ablation study comparing these architectures in Section~\ref{sec:ablation experiment}.

\subsubsection{Noise Injection into the MLLM.}
After obtaining the noise $\mathcal{E}$, the expression $q\left( A_i \mid X_{V_i}, X_{L_i}, \mathcal{E} \right)$ denotes the process of injecting the noise into the original MLLM architecture to assist visual question answering. Specifically, we consider both the injection position and the method of incorporating the noise. Since the noise shares the same shape as the input features, it is injected directly on top of the input features. Compared to approaches that use raw images and text as inputs to the noise generator, we leverage high-dimensional multimodal features extracted just before the LLM Decoder. 

This design offers several advantages: (1) these features are already preliminarily aligned through the pretrained model and contain \textbf{richer, more integrated semantic information}, which enhances the generator's ability to model complex semantics; (2) injecting noise closer to the output layer reduces the number of parameters involved in backpropagation, thereby \textbf{lowering training cost and improving efficiency}. For the injection method, we adopt an additive strategy with the visual features to preserve the original inference path as much as possible, minimizing architectural disruption and thereby \textbf{simplifying the training process}.

\section{Experiments}
We investigate various fine-tuning strategies—including full-parameter fine-tuning, LoRA, DoRA, and MBT—on two representative vision-language models: Qwen2.5-VL-3B/7B and LLaVA-1.5-7B. As presented in Section~\ref{sec:Main Results}, our Noise method surpasses these traditional fine-tuning approaches in accuracy while requiring fewer fine-tuned parameters than even LoRA. Moreover, through ablation studies (Section~\ref{sec:ablation experiment}) and noise visualization analysis (Section~\ref{sec:noise_vis}), we demonstrate the effectiveness of our proposed approach, which integrates cross-attention mechanisms with additive noise injection. This reveals the beneficial role of noise in guiding semantic understanding and enhancing model training in high-dimensional feature spaces.

\subsection{Datasets and Pre-trained Models}
    \textbf{Base Models and Fine-tuning Methods.}
    We conduct our experiments based on two representative vision-language models: LLaVA-1.5-7B~\cite{liu2023llava} and Qwen2.5-VL-3B/7B~\cite{Qwen2.5-VL}. We explore different fine-tuning strategies, including full-parameter fine-tuning, Low-Rank Adaptation (LoRA)~\cite{hu2022lora}, Weight-Decomposed Low-Rank Adaptation (DoRA)~\cite{liu2024dora}, High-rank updating(MoRA)~\cite{jiang2024mora}, and Multimodal Bottleneck Transformer (MBT)~\cite{nagrani2021attention} to evaluate the adaptability and efficiency of each method under various model scales and settings.
    
    \textbf{Fine-tuning Datasets.}
    We fine-tune our model using two high-quality multimodal datasets. LLaVA-Instruct-150K ~\cite{liu2023llava}] is a GPT-generated multimodal instruction-following dataset designed to improve visual instruction tuning and strengthen the alignment between visual understanding and language generation. MMPR-v1.1 ~\cite{wang2024enhancing, wang2024mpo, chen2024expanding, chen2024far, chen2023internvl} is a large-scale multimodal preference dataset containing about 3 million samples. It is specifically constructed to enhance models' reasoning abilities by optimizing for human-aligned responses in complex vision-language tasks.

 \subsection{Benchmarks and More Details}
 \textbf{Benchmarks.}
    We evaluate the fine-tuned models on widely adopted vision-language benchmarks: VQAv2~\cite{goyal2017making}, GQA~\cite{hudson2019gqa}, VisWiz~\cite{gurari2018vizwiz}, SQA~\cite{wang2024mpo}, TextVQA~\cite{singh2019towards}, POPE~\cite{li2023evaluating}, MMBench~\cite{liu2024mmbench}, MME~\cite{Fu2023MMEAC}, MM-Vet~\cite{Yu2023MMVetEL}, and ScienceQA~\cite{Lu2022LearnTE}. These benchmarks cover a wide range of vision-language capabilities, including natural image question answering, hallucination detection, multi-choice reasoning, and open-ended scientific or knowledge-intensive tasks.

\textbf{Training Details.}
To ensure fairness in comparison, we adopt the default configuration parameters provided in the official codebases of LLaVA and Qwen2-VL. As shown in Table~\ref{tab:training_settings_combined}, all models are fine-tuned using the same batch size and optimizer settings, while retaining the learning rates specified by their respective official implementations, on NVIDIA H100 GPUs.


\begin{table}[]
    \centering
    \setlength{\tabcolsep}{3pt}
    \begin{tabular}{l|cccc|cccc}
    \toprule
    \multirow{2}{*}{\textbf{Param.}}& \multicolumn{4}{c|}{\textbf{Qwen2.5VL-3B / 7B}} 
    & \multicolumn{4}{c}{\textbf{LLaVA-1.5-7B}} \\
    & FT.& Lo.& Do.& Mu.& FT.& Lo.& Do.& Mu.\\
    \midrule
    Batch Size & \multicolumn{4}{c|}{128} & \multicolumn{4}{c}{128} \\
    Learning Rate & 1e-5& 1e-5 & 1e-5 & 5e-4 & \multicolumn{4}{c}{2e-4} \\
    Warmup Rate& \multicolumn{4}{c|}{0.01} & \multicolumn{4}{c}{0.03} \\
    ZeRO Stage & Z-3& Z-2& Z-2& Z-3& Z-3& Z-3& Z-2& Z-2\\
    \bottomrule
    \end{tabular}
     \caption{Training settings of Full-FT, LoRA, DoRA, MBT  and MuNG on Qwen2.5VL-3B/7B and LLaVA-1.5-7B models. }
     \label{tab:training_settings_combined}
\end{table}

\label{sec:Main Results}
\begin{table*}[h]
    \centering 
    \footnotesize
    \setlength{\tabcolsep}{3pt} 
    \begin{tabular}{l@{\hspace{2pt}}r |c c c c c c ccc|c}
        \toprule
        \small \textbf{Method} &
        \small \textbf{\# Params.} &
        \small \textbf{MME-P} & 
        \small \textbf{MME-C} & 
        \small \textbf{MME(\%)} & 
        \small \textbf{SQA.} &
        \small \textbf{MMVet} &
        \small \textbf{MMStar} &
        \small \textbf{POPE} &
        \small \textbf{SEEDBench}& 
        \small \textbf{MMMU}&
        \small \textbf{Avg.} \\
        \midrule
        Base & \multicolumn{1}{c|}{-}       
        & 1592.40 & 607.50 & 78.57 
        & 79.30 & 60.00 & 56.30 &    85.90 
&74.00 &- & -\\
        Base* & \multicolumn{1}{c|}{-}        
        & 1563.00 & 584.00 & 76.68 
        & 79.68 & 65.00 & 54.30 &  86.32 
&73.50 & 47.33 & 68.97 \\
        \specialrule{0.1pt}{0pt}{0.4pt}
        \specialrule{0.1pt}{0.4pt}{0pt}
        Full-FT* & 100.00\%  
        & 1555.21 & 587.14 & 76.51 
        & \textbf{80.68} & \underline{66.20} & 43.26 &    85.50 
&71.93 
& 52.00 & 68.01 
\\
        MBT*& $<$0.01\%
         & 1444.54  & 506.43  & 69.68  
        & 76.06  & 50.90  & 51.00  &    84.82  
&71.69 
& 38.00 
& 63.16 
\\
        LoRA*& 7.82\%
         & \textbf{1624.15} & 612.50 & \underline{79.88} 
        & 79.25 & 65.30 & \textbf{55.33} &    86.50 
&\textbf{73.29} & \textbf{53.33} & \underline{70.41 }
\\
        MoRA* & 0.78\%
        & 1579.94 & 614.64 & 78.38 
        & 77.20 & 52.30 & 53.00 &    \textbf{88.50} 
&71.98 & 48.00 & 67.05 
\\
        DoRA* & 7.99\%
        & 1566.64 & \textbf{638.93} & 78.77 
        & 79.25 & 65.40 & 54.06 &    86.37 
&73.41 & 48.67 & 69.42 
\\
        \rowcolor{Gray}
        MuNG(Ours)* & 0.67\%
        & \underline{1612.66} & \underline{625.00} & \textbf{79.92} 
        & \underline{79.54} & \textbf{66.50} & \underline{54.46} &   \underline{86.95} &\textbf{73.64} & \textbf{53.33} & \textbf{70.62} 
\\
        \bottomrule
    \end{tabular}
    \caption{\textbf{Comparsion with other methods on Qwen2.5-VL-3B.} We perform visual instruction tuning on the MMPR-v1.1 dataset. Our method achieves the best results on most benchmarks and obtains the highest average score overall.  \textbf{Base} results are from the official Qwen release (Bai et al., 2025). \textbf{$*$} Indicates evaluation using the same prompts and LLM evaluator ~\cite{duan2024vlmevalkit}.}
    \label{tab:qwen3B_Result}
\end{table*}

\begin{table*}[h]
    \centering
    \begin{tabular}{l r |c c c  c c |c}
    \toprule
    \small \textbf{\textbf{Method}} &
    \small \textbf{\textbf{\# Params.}} &
    \small \textbf{MME-P} & 
    \small \textbf{MME-C} & 
    \small \textbf{Sum(\%)} &
    \small \textbf{MMVet} &
    \small \textbf{POPE} &
    \small \textbf{Avg.} \\
    \midrule
    Base & \multicolumn{1}{c|}{-}          
    & 1698.10 & 613.90 & 82.57 & 69.70 & 85.90 & 79.39 
\\
    
    Base* & \multicolumn{1}{c|}{-}          
    & 1693.53 & 611.43 & 82.32 & 72.00& 
87.01& 80.44 
\\
    \specialrule{0.1pt}{0pt}{0.4pt}
    \specialrule{0.1pt}{0.4pt}{0pt}
    Full-FT* & 100.00\%  
    & \underline{1692.79} & \underline{630.71} & \underline{82.98} & 69.00 & \underline{87.28} & 79.75 
\\
    MBT* & $<$0.01\%& 1609.40 & 512.86 & 75.79 & 61.00 & 85.38 & 74.06
        \\
    LoRA* & 6.38\%& 1645.74 & 626.79 & 81.16 & \textbf{72.20} & 86.74 & \underline{80.03} 
\\
    MoRA* & 0.48\% & 1690.83 & 620.00 & 82.53 & 52.30 & 87.22 & 74.00 
\\
    DoRA* & 6.38\% & 1580.88 & \textbf{638.93} & 79.28 & 66.60 & 86.29 & 77.39 
\\
    \rowcolor{Gray}
    MuNG (Ours)* & 1.83\% & \textbf{1717.34} &609.64 & \textbf{83.11} & \underline{71.00} &\textbf{87.41} & \textbf{80.51} 
\\
    \bottomrule
  \end{tabular}
  \caption{\textbf{Comparsion with other methods on Qwen2.5-VL-7B.}  We perform visual instruction tuning on the MMPR-v1.1 dataset. Our method achieves the best results on most benchmarks and obtains the highest average score overall. The \textbf{Base} results are from the official release by the Qwen team. The \textbf{Base*} results are obtained using the Qwen2.5-VL-7B-Instruct parameters, evaluated with the same prompts and evaluation LLM as other methods, using the VLMEvalKit toolkit.}
      \label{tab:qwen7B_Result}
\end{table*}

\textbf{Eval Toolkits.}
For evaluation, we use the official testing scripts from both the LLaVA and DoRA repositories for LLaVA models. For Qwen2.5-VL models, we adopt the VLMEvalKit toolkit ~\cite{duan2024vlmevalkit}, ensuring consistency in prompts and scoring by using the same evaluation LLM, specifically InternLM2.5-1.8B-Chat ~\cite{cai2024internlm2}.
It is worth noting that due to differences in third-party Python library versions, prompt selection, and other environmental factors, we were unable to perfectly reproduce leaderboard results. Therefore, we present both the leaderboard-reported scores and our reproduced results under the same evaluation conditions for comparison and analysis.


\begin{figure*}[h]
  \centering
  \makebox[\linewidth]{\includegraphics[width=1\textwidth]{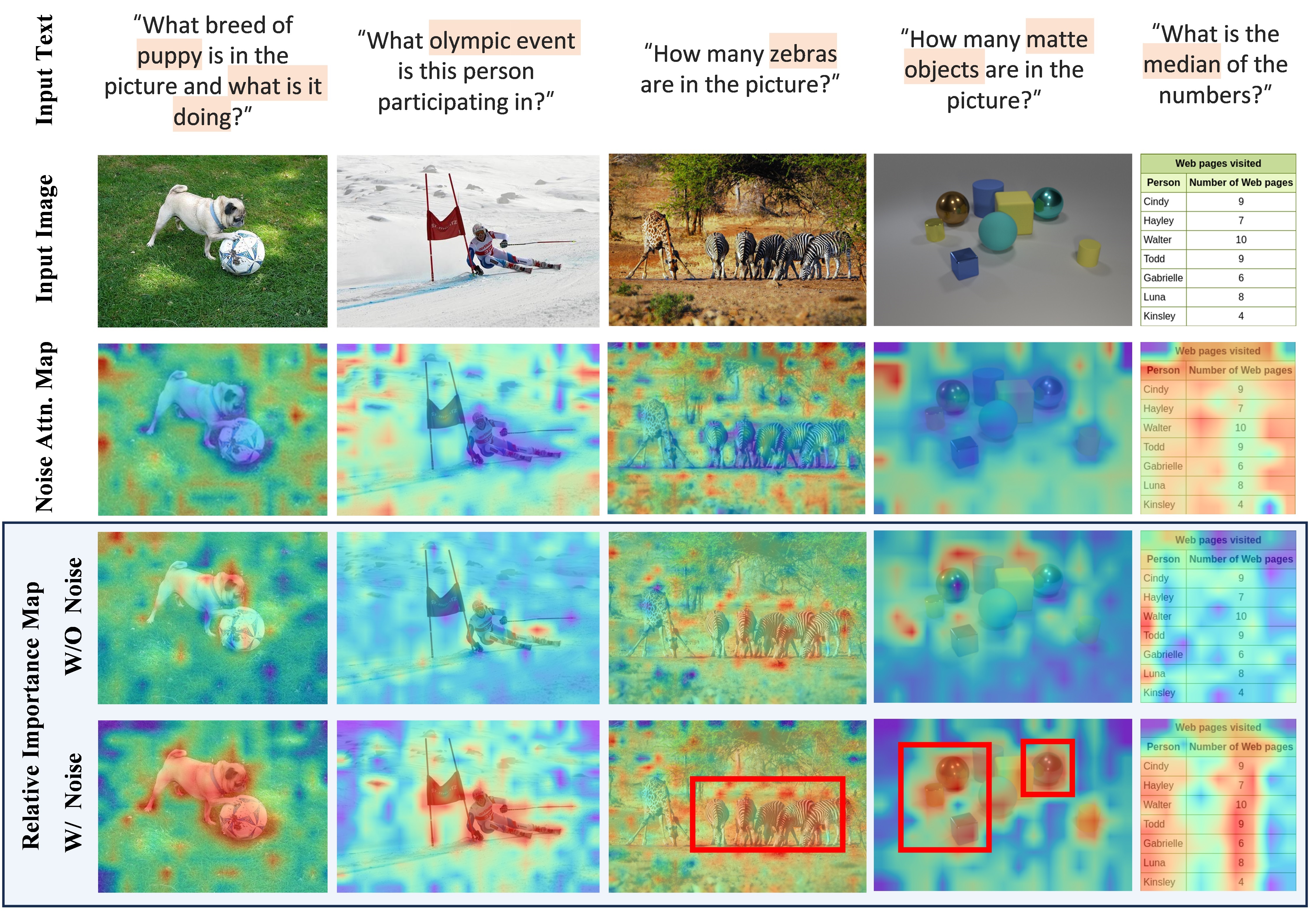}}
  \caption{\textbf{ Visualization of the generated noise injected into high-dimensional visual features. }
    The top three rows show the input text, images, and noise module's attention maps; the bottom two show visual-text importance maps before and after noise injection.
    The attention maps indicate that MuNG can effectively identify and selectively \textbf{suppress} semantically irrelevant or unmentioned regions in the image. The relative importance maps further highlight that the noise enhances the representation of image regions that are more crucial for answering the question.
  }
  \label{fig:noise_visualization}
\end{figure*}

\begin{table}[t]
  \centering
  \footnotesize
  \setlength{\tabcolsep}{0.5pt}
  \begin{tabular}{l c|c c c c c c |c}
    \toprule
    \textbf{Method }& \textbf{\# Param.} & \textbf{GQA}& \textbf{VisWiz}& \textbf{SQA}& \textbf{T-VQA}& \textbf{POPE}& \textbf{MMVet}& \textbf{Avg.}
\\
    \midrule
    Full-FT & 100.00\%   & 61.9 & \underline{50.0} & 67.2 & \textbf{58.2} & 85.9 & 31.1 & 59.1 
\\
    LoRA & 4.61\%  & \textbf{62.9} & 47.8 & 68.3 & \textbf{58.2} & 86.4 & 30.2 & 59.0 
\\
    DoRA &  4.63\%  & \textbf{62.9} & \textbf{52.2} & \underline{68.4} & 57.0 & \textbf{87.2} & \textbf{33.3} & \textbf{60.2} 
\\
    \rowcolor{Gray}
    MuNG & 2.78\%& 61.5 & 49.8 & \textbf{70.0} & 55.4 & \underline{86.9} & \underline{32.4} &\underline{59.3} 
\\
    \bottomrule
  \end{tabular}
  \caption{\textbf{Comparsion with other methods on LLaVA-1.5-7B.} We directly use the publicly available checkpoints of LLaVA ~\cite{liu2023llava} and DoRA ~\cite{liu2024dora} to reproduce their results. }
   \label{tab:llava_results}
\end{table}

\subsection{Main Results}
\subsubsection{Evaluation of Fine-Tuning Methods on Qwen}
The Qwen series adopts the latest Qwen2.5-VL-3B/7B-Instruct~\cite{Qwen2.5-VL}, and is partially fine-tuned using a subset of the MMPR-V1.1~\cite{wang2024enhancing} dataset. As shown in Table~\ref{tab:qwen3B_Result}, Table~\ref{tab:qwen7B_Result}.
 We conduct evaluations under the same settings, using same prompts and LLM evaluator ~\cite{duan2024vlmevalkit}. Our method achieves an average accuracy across tasks that surpasses Full-FT, suggesting that Full-FT may suffer from overfitting during training. Furthermore, MuNG consistently outperforms LoRA, DoRA, MBT, and Full-FT in overall performance.


 Specifically, on Qwen2.5-VL-3B, our method consistently surpasses Full-FT, MBT, LoRA, MoRA, and DoRA in average accuracy, while requiring only a small fraction of trainable parameters. A similar trend holds on Qwen2.5-VL-7B, where our method again outperforms the aforementioned approaches, yet still tunes far fewer parameters than LoRA, DoRA, and Full-FT.

These results indicate that MuNG not only delivers competitive performance but also offers significant advantages in parameter efficiency, highlighting its potential as an effective and compact fine-tuning approach.

\subsubsection{Evaluation of Fine-Tuning Methods on LLaVA}

As shown in Table~\ref{tab:llava_results}, our method MuNG demonstrates competitive performance compared to other fine-tuning approaches on the LLaVA-1.5-7B model. 
It is important to note that our fine-tuning on LLaVA-1.5-7B is based on the pretrained model LLaVA-v1.5-7B-pretrain. During the first-stage pretraining of LLaVA, only the multimodal alignment layer was trained on multimodal alignment data, while the LLM decoder was directly adopted from a pretrained LLM and trained solely on textual data.

If all components—both the modality encoders and the LLM decoder—remain frozen during fine-tuning and only the parameters of MuNG are updated, its performance is significantly constrained. To address this issue, we propose a simple yet effective solution: we freeze the original model and introduce a small number of low-rank LoRA adapters, which are fine-tuned jointly with MuNG. This approach substantially restores MuNG’s performance and further demonstrates the necessity of re-tuning the LLM decoder in multimodal large language models (MLLMs). Additionally, our method requires far fewer trainable parameters compared to standard LoRA-based approaches that achieve similar results. More ablation studies can be found in the appendix.

For fine-tuning, we adopt the Instruct dataset provided by LLaVA. In terms of parameter efficiency, LoRA and DoRA update approximately $4.6\%$. In contrast, MuNG updates only $2.78\%$ of the parameters. Despite its minimal parameter footprint, MuNG achieves the highest score on ScienceQA ($70.0\%$), and delivers near-optimal performance on POPE and MM-Vet, consistently ranking among the top. Overall, MuNG drastically reduces the number of trainable parameters while maintaining or surpassing the performance of more parameter-intensive baselines.

\subsection{Ablation Study}
\label{sec:ablation experiment}
\begin{table}[h]
  \centering
  \setlength{\tabcolsep}{1.5pt}
  \begin{tabular}{c c c |c c cc|c}
    \toprule
    Struct.& MM.& Noise
& MME& SQA& MMStar&SEEDBench
& \textbf{Avg.} \\
    \midrule
    MLP& add& w/& 33.70 & 69.52 & 28.07 &40.11 
& 42.85 
\\
    MLP& dot& w/& 27.92 & 69.00 & 26.40 &41.11 
& 41.11 
\\
    CA& dot& w/& 32.38 & 67.57 & 28.20 &52.45 
& 45.15 
\\
    \specialrule{0.2pt}{0pt}{0.4pt}
 CA& add& w/o& 76.69 & 78.44 & 54.07 &72.78 
&70.49 
\\
 Gauss.& add& w/& 79.10& 79.16
& 54.27 &73.41 
&71.48 
\\
\rowcolor{Gray}
    CA& add& w/& \textbf{79.92} & \textbf{79.54} & \textbf{54.46} &\textbf{73.64} 
& \textbf{71.89} 
\\
    \bottomrule
  \end{tabular}
   \caption{We analyze the model performance under different injection strategies and architectural designs: MM. means Merge Method; Add refers to additive injection; Dot denotes multiplicative (dot-product) injection; CA stands for Cross-Attention architecture.}
     \label{tab:Abblations}
\end{table}

 As shown in Table~\ref{tab:Abblations}, we analyze the impact of different injection strategies and architectural designs on model performance. Among these variants, the method combining CA+Add achieves the best performance. 
 
 Furthermore, we examine a variant where the Cross-Attention module is used solely for feature extraction, and its output is added to the original features—thus preserving the cross-attention structure but without incorporating beneficial noise sampling. This design allows us to isolate the contribution of the CA mechanism itself.
 We also attempted to directly add Gaussian noise to the image features to explore the role of randomness. This result highlights that the key factor is not the randomness of the noise, but the informative guidance it provides to the model.

The results show that the performance gain does not simply arise from using a CA structure for cross-modal alignment. Instead, it suggests that the key factor lies in the integration of useful noise representations, rather than the CA module alone.

\subsection{Noise visualisation and analysis}
\label{sec:noise_vis}
To further understand the role of noise in MLLMs, we visualize the generated noise. As shown in Figure~\ref{fig:noise_visualization}, we select examples from various visual question answering tasks to analyze the distribution characteristics of the injected noise. 
The attention maps in the third row demonstrate that the proposed MuNG effectively identifies semantically relevant regions within the image. The more blue an area appears, the less likely MuNG is to apply a mask, indicating that the region is important for the MLLM to attend to. 
The fourth and fifth rows show visual-text feature importance maps before and after noise injection. For instance, in the first and second columns, MuNG accurately suppresses regions irrelevant to the input text, such as grass and snowfields. 

Notably, MuNG not only masks non-central background regions but also selectively suppresses salient objects that are not mentioned in the text. 
For example, in the third column, MuNG shows different attention patterns between the giraffe and the queried zebra; similarly, in the fourth column, it distinguishes between the unmentioned matte objects and other relevant elements.

Although the generated noise is not directly interpretable by the human eye, it introduces meaningful perturbations in the high-dimensional feature space, providing stronger semantic guidance and enhanced discriminative capacity during training. 
Additional visualizations of the noise mean and variance are provided in the appendix.

\subsection{Efficiency evaluation}

\begin{table}[ht]
    \centering
    \setlength{\tabcolsep}{3.5pt} 
    \begin{tabular}{lccccc}
        \toprule
        Method & \begin{tabular}{c} \#Train. \\(\%)\end{tabular} & \begin{tabular}{c}Training \\Time ($\times$)\end{tabular} &\   \begin{tabular}{c}TTFT \\(s)\end{tabular} &
        \begin{tabular}{c}TPOT \\(µs)\end{tabular} &
        \begin{tabular}{c}Avg.\end{tabular} \\
        \midrule
        FT    & 100.00\%  & 5.17$\times$ &   \textbf{0.9} 
&\textbf{20.5}&68.01 
\\
 MBT& \textbf{$<$0.01\%} & 1.33$\times$& 2.6 
& 22.0 
& 63.16\\
        LoRA  & 7.99\% & 2.42$\times$ &   2.5 
&21.4&\underline{70.41 }
\\
        DoRA  & 7.82\% & 1.58$\times$ &   \underline{2.0} 
&21.7&67.05 
\\
        MoRA  & 0.78\% & \underline{1.25}$\times$ &   5.2 
&20.6&69.42 
\\
        \rowcolor{Gray}
        MuNG  & \underline{0.67\%} & \textbf{1.00$\times$} &   3.2 
&\textbf{20.5}&\textbf{70.62} 
\\
        \bottomrule
    \end{tabular}
    \caption{ Efficiency evaluation and performance com
parison. Results are based on
 Qwen2.5-VL-3B.}
 \label{tab:Efficiency}
\end{table}

In terms of fine-tuning efficiency, we further evaluate the proposed MuNG method and compare it with FT, MBT, LoRA, DoRA, and MoRA baselines in terms of the proportion of trainable parameters relative to the full model and the overall training time. As shown in Table~\ref{tab:Efficiency}, all these parameter-efficient fine-tuning methods introduce only a small number of additional parameters compared to full MLLMs. Moreover, MuNG significantly reduces training time, while achieving consistent performance improvements across multiple benchmarks. And MuNG has slightly longer TTFT but matches the best TPOT. In summary, MuNG achieves notable performance gains with only a small increase in trainable parameters, demonstrating its effectiveness, efficiency, and parameter-efficient, fast-to-train design with minimal impact on inference speed.

\section{Conclusion}
\label{sec:Conclusion}
We reformulate MLLM inference via variational inference and propose MuNG, which generates beneficial noise from multimodal data to suppress irrelevant semantics and improve cross-modal representations. With only about $1\%$ extra parameters, MuNG outperforms full fine-tuning, LoRA, DoRA, and MBT, offering an efficient adaptation method. Future work will explore its use across more modalities.

Although our method offers many advantages, it is not without limitations. When the LLM Decoder has not been pretrained on multimodal data at all, keeping all modal Encoders and the LLM Decoder frozen while only fine-tuning MuNG leads to a noticeable performance drop. To address this issue, we propose a simple yet effective solution: while keeping most parameters fixed, we introduce a small low-rank LoRA Adapter to fine-tune the LLM Decoder. This allows MuNG’s performance to be substantially restored, while the total number of trainable parameters remains significantly lower than that required by pure LoRA approaches achieving similar performance. Our experiments validate the effectiveness of this strategy. 

\bibliography{aaai2026}

\newpage








\title{Explore How to Inject Beneficial Noise in MLLMs}

%

\setcounter{secnumdepth}{2}

\author{%
  David S.~Hippocampus\thanks{Use footnote for providing further information
    about author (webpage, alternative address)---\emph{not} for acknowledging
    funding agencies.} \\
  Department of Computer Science\\
  Cranberry-Lemon University\\
  Pittsburgh, PA 15213 \\
  \texttt{hippo@cs.cranberry-lemon.edu} \\
}



\appendix

\section{Technical Appendices and Supplementary Material}

\subsection{Additional Ablation Studies}


\begin{table}[h]
  \centering
  \footnotesize
  \setlength{\tabcolsep}{2.5pt}
  \begin{tabular}{l @{\hspace{5pt}}c|c c c |c}
    \toprule
    \textbf{Method} & \textbf{Rank}
& \textbf{SQA-IMG} & \textbf{POPE} & \textbf{MM-Vet}& \textbf{Avg.}
\\
    \midrule
    Full-FT & -
& 67.2 & 85.9 & \underline{31.1} & 61.4\\
    LoRA & 128
& 68.3 & 86.4 & 30.2 & 61.6\\
    LoRA & 32
& \underline{68.4} & \underline{86.8} & 30.5 
& \underline{61.9}\\
    \rowcolor{Gray}
    LoRA+MuNG& 32& \textbf{70.0} & \textbf{86.9} & \textbf{32.4} &\textbf{63.1}\\
    \bottomrule
  \end{tabular}
  \caption{\textbf{Ablation Study on the Rank of LoRA Adapters}}
  \label{tab:llava_results}
\end{table}

Our fine-tuning on the LLaVA-1.5-7B model is based on the pretrained model LLaVA-v1.5-7B-pretrain. This model is obtained from the first-stage pretraining of LLaVA, during which only the multimodal alignment layer is trained on multimodal data, while the language decoder (LLM)  is directly adopted from a pretrained Vicuna-7B-v1.5 model. This LLM decoder is trained exclusively on textual data and has not been exposed to any multimodal inputs.

If all components—including the modality encoder and the LLM decoder—remain frozen during fine-tuning, with only the parameters of MuNG being updated, the model's performance is significantly constrained. This effectively replicates the setup of the first stage of LLaVA and highlights the limitation of directly using a frozen pretrained LLM in multimodal large language models (MLLMs). It further confirms that fine-tuning the LLM decoder is essential for achieving strong performance.

To address this issue, we propose a concise yet effective strategy: we freeze the original model and introduce only a small number of low-rank LoRA adapters, which are jointly fine-tuned together with MuNG. As shown in Table~\ref{tab:llava_results}, we compare the results of using LoRA adapters with different ranks and find that combining MuNG with low-rank LoRA adapters can effectively improve overall accuracy.

In summary, our lightweight fine-tuning approach demonstrates that introducing a small number of updates to a frozen LLM for adaptation to multimodal tasks is not only feasible but also beneficial—especially when combined with a structured module like MuNG, which is designed to mask semantically irrelevant content.

\subsection{Additional Results}
\label{sec:noise_vis}

\begin{figure*}[h]
  \centering
  \makebox[\linewidth]{\includegraphics[width=1\textwidth]{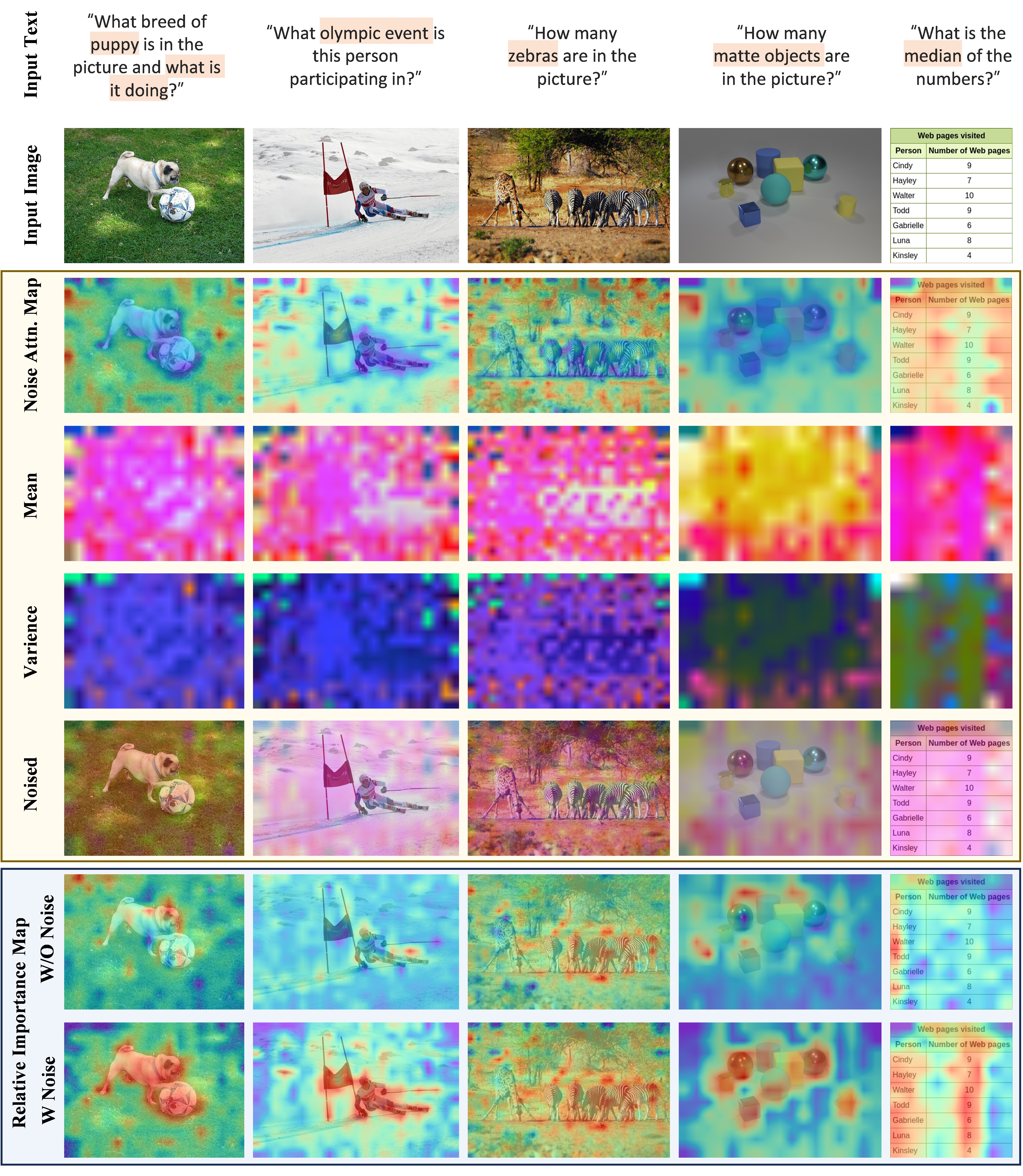}}
  \caption{ Visualization of generated noise injected into high-dimensional visual features.}
  \label{fig:noise_vis_app}
\end{figure*}

We visualize the noise injected into the original images using various types of Visual Question Answering data, as shown in Figure~\ref{fig:noise_vis_app}. The first row displays the input text, the second row shows the original input images, and the third row presents the attention maps generated by the cross-attention-based noise module. The fourth and fifth rows correspond to the mean and variance of the noise, respectively, which are reduced to 3 dimensions via PCA for visualization. The fifth row also illustrates the effect of adding the noise onto the original image. The bottom two show visual-text importance maps before and after noise injection.

Particular attention should be paid to the third column, where the question asks about the number of zebras in the image. From the visualizations of the mean and variance, it is evident that MuNG effectively distinguishes between the unmentioned giraffes and the target zebras.

Through this visualization, we further emphasize that MuNG is not merely a data augmentation technique. It demonstrates strong semantic understanding capabilities, enabling it not only to distinguish between foreground objects and background regions but also to identify semantically relevant information crucial for answering the given questions—thereby simplifying the overall task. This aligns well with our design principle of reducing task entropy.

\end{document}